\documentclass[10pt,twocolumn,letterpaper]{article}

\usepackage{iccv}
\usepackage{times}
\usepackage{epsfig}
\usepackage{graphicx}
\usepackage{amsmath}
\usepackage{amssymb}

\usepackage{url}
\usepackage{comment}
\usepackage{color}
\usepackage{algorithm}
\usepackage{algpseudocode}
\usepackage{subfigure}
\usepackage{enumitem}
\usepackage{booktabs}

\makeatletter
\newcommand{\algmargin}{\the\ALG@thistlm}
\makeatother
\newlength{\whilewidth}
\algdef{SE}[parWHILE]{parWhile}{EndparWhile}[1]
{\parbox[t]{\dimexpr\linewidth-\algmargin}{%
		\hangindent\whilewidth\strut\algorithmicwhile\ #1\ \algorithmicdo\strut}}{\algorithmicend\ \algorithmicwhile}
\algnewcommand{\parState}[1]{\State \parbox[t]{\dimexpr\linewidth-\algmargin}{\strut #1\strut}}

\usepackage{bm}
\usepackage{multirow}
\usepackage{wrapfig}
\usepackage{xcolor}
\usepackage{color, colortbl}
\definecolor{Gray}{gray}{0.9}
\definecolor{White}{rgb}{1,1,1}
\usepackage[english]{babel}
\usepackage[breaklinks=true,bookmarks=false]{hyperref}

\iccvfinalcopy 

\def\iccvPaperID{****} 

\ificcvfinal\fi

\begin{document}

\title{Dynamic Divide-and-Conquer Adversarial Training for \\Robust Semantic Segmentation}

\author{Xiaogang Xu$^1$ \quad Hengshuang Zhao$^{2,3}$ \quad Jiaya Jia$^{1,4}$\\
$^1$ The Chinese University of Hong Kong \quad $^2$ University of Oxford \\$^3$The University of Hong Kong \quad $^4$ SmartMore\\
{\tt \small \{xgxu, leojia\}@cse.cuhk.edu.hk, \quad hszhao@cs.hku.hk}
}

\maketitle
\ificcvfinal\thispagestyle{empty}\fi

\begin{abstract}
Adversarial training is promising for improving robustness of deep neural networks towards adversarial perturbations, especially on the classification task. The effect of this type of training on semantic segmentation, contrarily, just commences. We make the initial attempt to explore the defense strategy on semantic segmentation by formulating a general adversarial training procedure that can perform decently on both adversarial and clean samples. We propose a dynamic divide-and-conquer adversarial training (DDC-AT) strategy to enhance the defense effect, by setting additional branches in the target model during training, and dealing with pixels with diverse properties towards adversarial perturbation. Our dynamical division mechanism divides pixels into multiple branches automatically. Note all these additional branches can be abandoned during inference and thus leave no extra parameter and computation cost. Extensive experiments with various segmentation models are conducted on PASCAL VOC 2012 and Cityscapes datasets, in which DDC-AT yields satisfying performance under both white- and black-box attack.
The code is available at \href{https://github.com/dvlab-research/Robust-Semantic-Segmentation}{https://github.com/dvlab-research/Robust-Semantic-Segmentation}.
\end{abstract}

\section{Introduction}
\begin{figure}[t]
	\centering
	\small
	\begin{tabular}{@{\hspace{0.0mm}}c@{\hspace{1.0mm}}c@{\hspace{1.0mm}}c@{\hspace{0.0mm}}}
		\includegraphics[width=0.32\linewidth, height=0.16\linewidth]{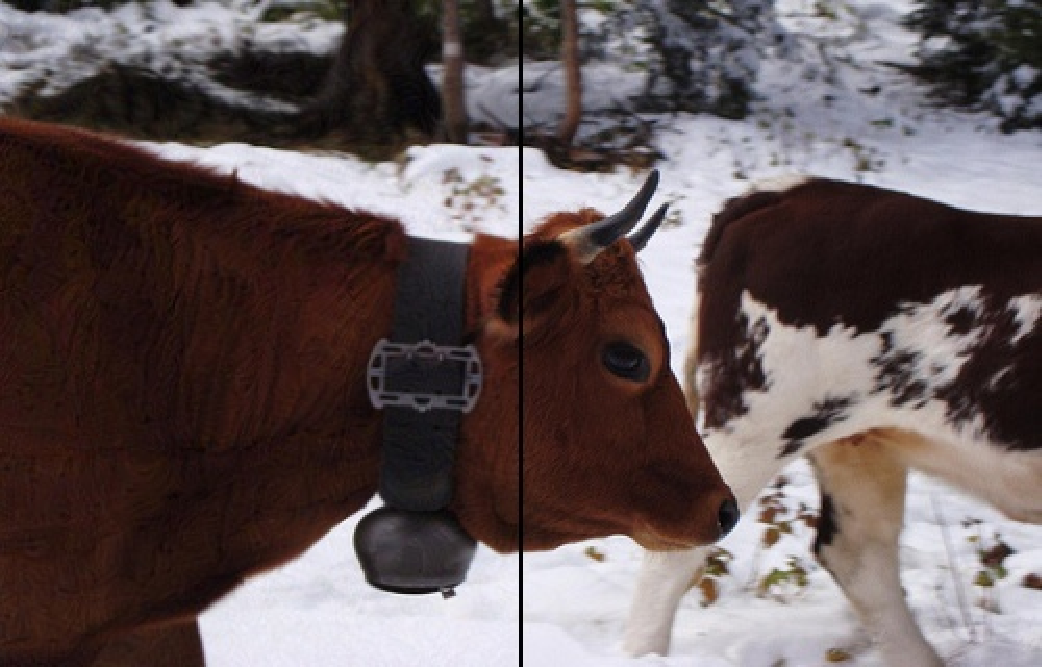}&
		\includegraphics[width=0.32\linewidth, height=0.16\linewidth]{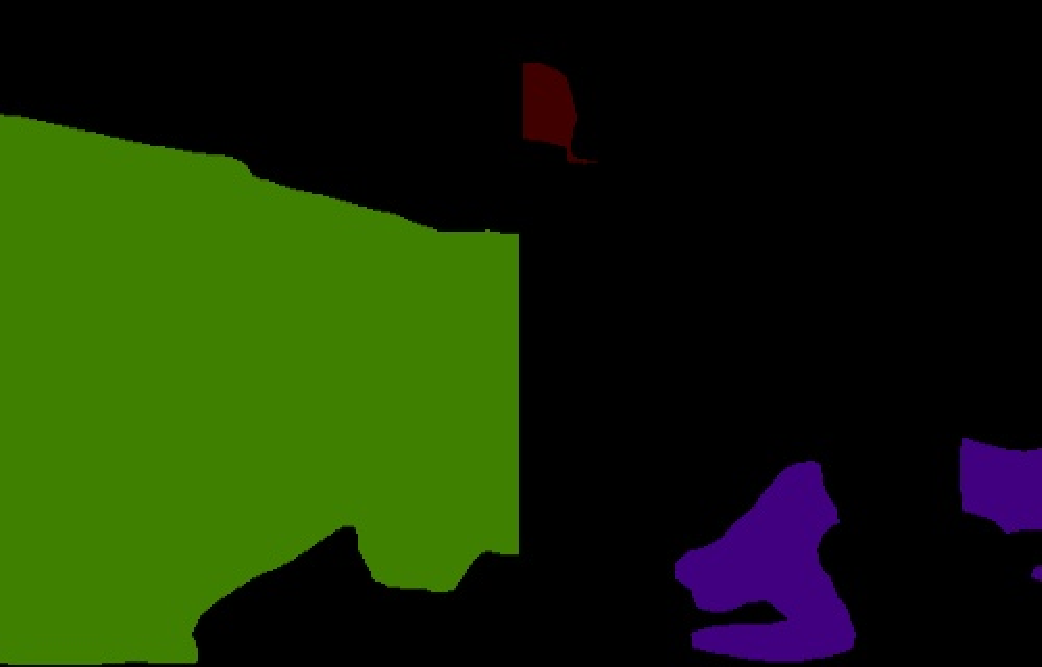}&
		\includegraphics[width=0.32\linewidth, height=0.16\linewidth]{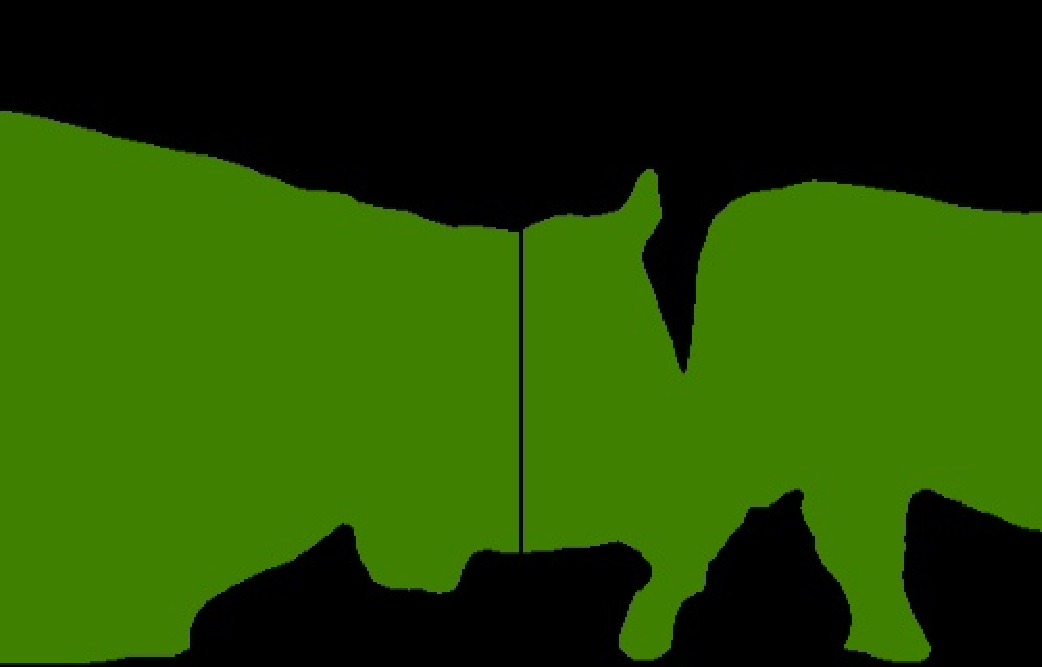}\\
		\includegraphics[width=0.32\linewidth, height=0.16\linewidth]{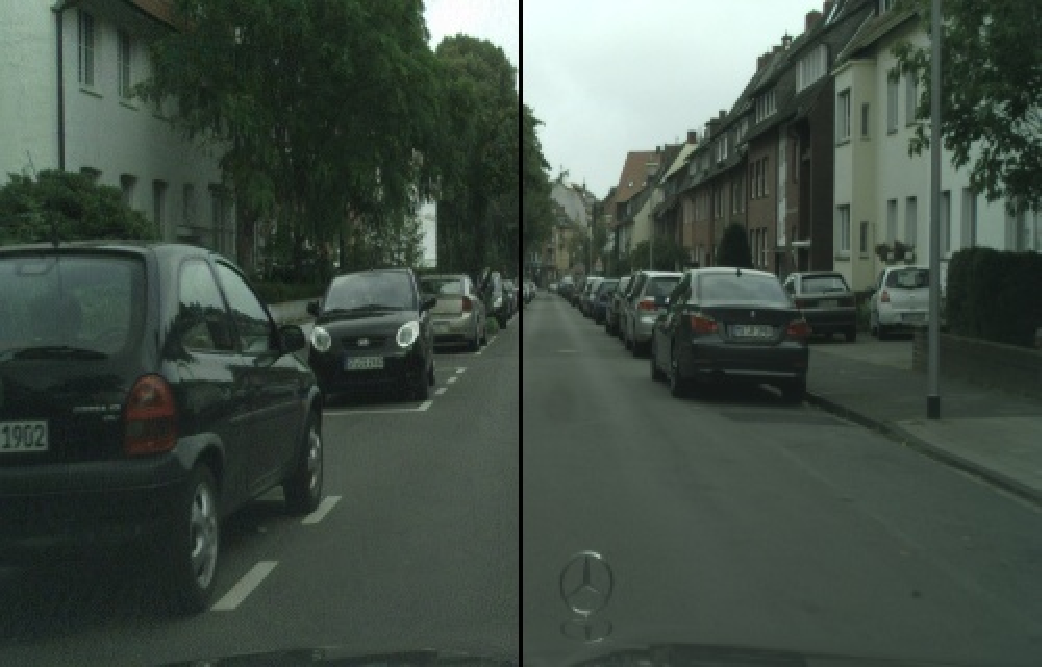}&
		\includegraphics[width=0.32\linewidth,height=0.16\linewidth]{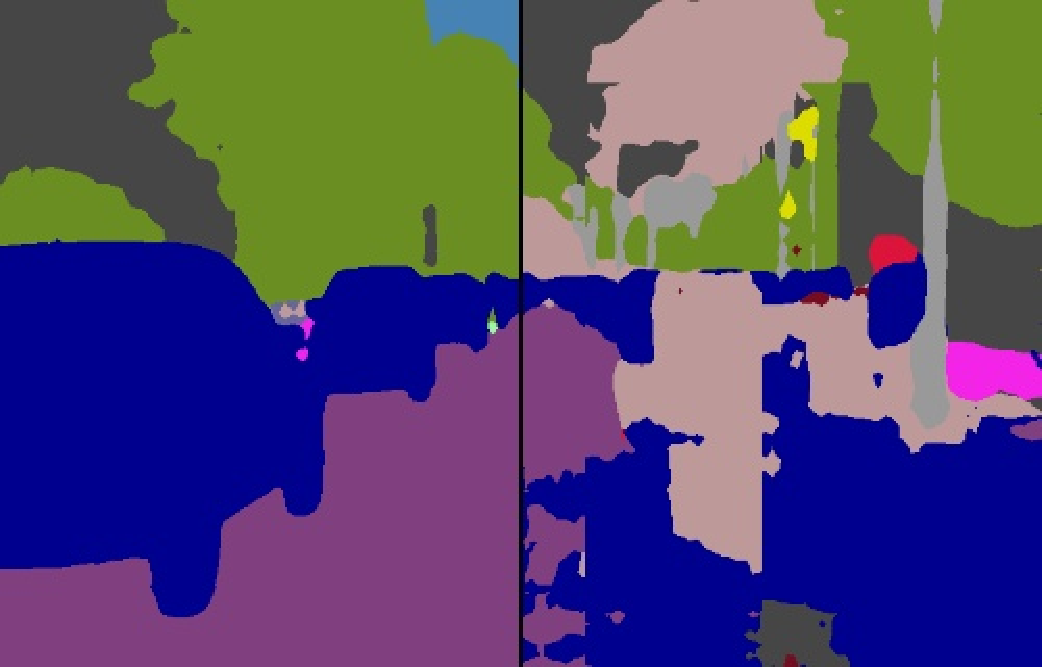}&
		\includegraphics[width=0.32\linewidth,height=0.16\linewidth]{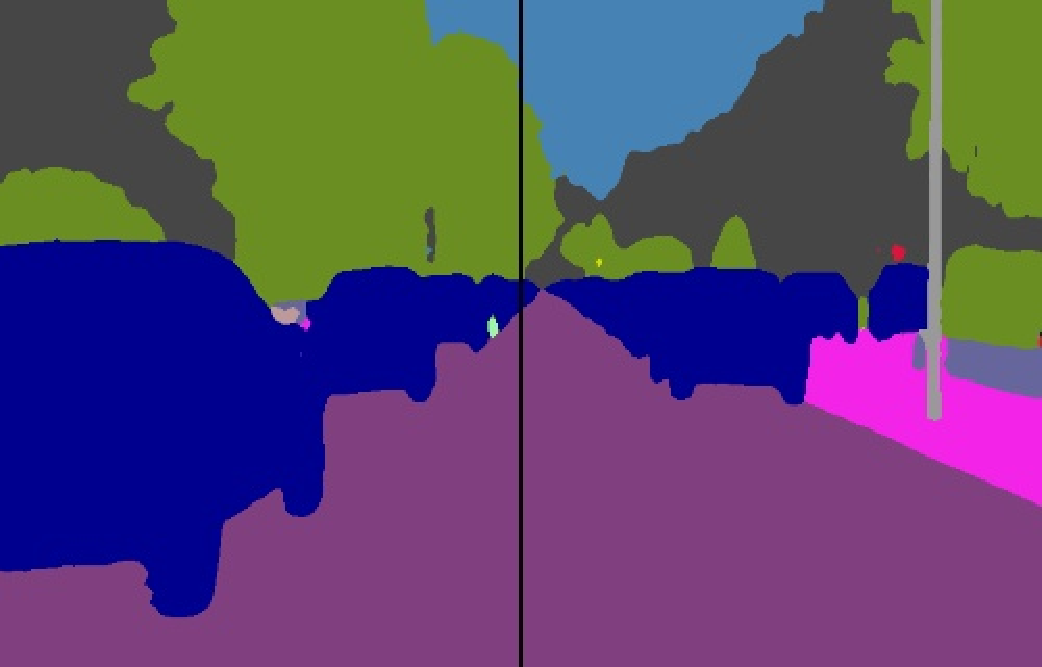}\\
		(a) Image & (b) No Defense & (c) With Our Defense \\
	\end{tabular}
	\caption{For each image in (a), the left side is the normal data while the right side is perturbed by adversarial noise. (b) shows that the adversarial attack could fail existing segmentation models. We provide an effective defense strategy shown in (c). The top and bottom rows are results with PSPNet and DeepLabv3, respectively.}
	\vspace{-0.2in}
	\label{fig:short}
\end{figure}

Recent work has revealed that deep learning models, especially in the classification task, are often vulnerable to adversarial samples~\cite{szegedy2013intriguing,goodfellow2014explaining,papernot2016limitations}. The adversarial attack can deceive the target model by generating crafted adversarial perturbations on original clean samples. Such perturbations are often imperceptible. Meanwhile, such threat also exists in semantic segmentation~\cite{xie2017adversarial,metzen2017universal,arnab2018robustness}, as shown in Fig. \ref{fig:short}.
However, there is seldom work to improve the robustness of semantic segmentation networks. As a universal approach, \textit{adversarial training} \cite{goodfellow2014explaining,kurakin2016adversarial2,madry2017towards} is effective to enhance the target model in classification by training models with adversarial samples. In this paper, we study the effect of adversarial training on the semantic segmentation task. We find that adversarial training impedes convergence on clean samples, which also happens in classification. 
We aim to make networks perform well on adversarial examples and meanwhile maintaining good performance on clean samples.

For the semantic segmentation task, each pixel has one classification output. Thus the property of every pixel in one image toward adversarial perturbations might be different.
Based on this motivation, we design a dynamic divide-and-conquer adversarial training (DDC-AT) strategy. We propose to use multiple branches in the target model during training, each handling pixels with a set of properties. During training, a ``main branch'' is adopted to deal with pixels from adversarial samples and pixels from clean samples that are not likely to be perturbed; an ``auxiliary branch'' is utilized to deal with pixels from clean samples that are sensitive to perturbations.

Moreover, such a divide-and-conquer setting is dynamic.
During training, pixels near the decision boundary from clean samples are initially set to the ``auxiliary branch''. They become more insensitive to perturbations in the ``auxiliary branch'', and finally move back to the ``main branch'' for processing. 
Such a dynamic procedure is implemented by training a ``mask branch''.
With this mechanism, our method reduces performance deterioration over clean samples. Experiments manifest that such a mechanism also improves robustness towards adversarial samples.
Another notable advantage of our proposed DDC-AT is that branches apart from the main one can be abandoned during inference. Thus parameters and computation cost remain almost the same.
We conduct extensive experiments with various segmentation models on both PASCAL VOC 2012~\cite{everingham2010pascal} and Cityscapes~\cite{cordts2016cityscapes} datasets. It is validated that our standard adversarial training strategy is effective to improve the robustness of segmentation networks, and our new DDC-AT strategy further boosts the effect of defense. It yields superior performance under both white- and black-box attacks. 

In summary, our main contribution is threefold.
\vspace{-0.05in}
\begin{itemize}
	\setlength{\itemsep}{1pt}
	\setlength{\parsep}{1pt}
	\setlength{\parskip}{1pt}
	
	\item It is the first attempt to have a comprehensive exploration on the effect of adversarial training for semantic segmentation. Our standard adversarial training can be treated as a strong baseline to evaluate defense strategies for semantic segmentation networks.
	\item We propose the DDC-AT to notably improve the defense performance of segmentation networks on both clean and adversarial samples.
	\item We conduct experiments with various model structures on different datasets, which manifest the effectiveness and generality of DDC-AT.
\end{itemize}

\vspace{-0.05in}
\section{Related Work}
\vspace{-0.05in}
\noindent{\textbf{Adversarial attack.}} The adversarial attack can be divided into two categories of white-box attack \cite{athalye2018robustness,goodfellow2014explaining}, where attackers have complete knowledge of the target model, and black-box attack \cite{papernot2017practical,papernot2016transferability,inkawhich2019feature,wu2020boosting}, where attackers have almost no knowledge of the target model.
Existing adversarial attack methods focus on solving the image classification problem. Such attack is normally achieved by computing or simulating the gradient information of target models \cite{goodfellow2014explaining,tramer2017ensemble,dong2018boosting,kurakin2016adversarial}. Meanwhile, as indicated by several recent methods \cite{xie2017adversarial,metzen2017universal,arnab2018robustness}, semantic segmentation networks are also vulnerable to adversarial samples.

\noindent{\textbf{Adversarial defense.}} Current defense methods for the classification task can be divided into four kinds: 1) changing the input of networks to remove perturbation \cite{jia2019comdefend,guo2017countering,prakash2018deflecting,song2017pixeldefend}; 2) adopting random strategy to obtain correct output \cite{xie2017mitigating,raff2019barrage,ding2019defending,dhillon2018stochastic}; 3) designing robust structures for different tasks \cite{xie2019feature,gu2014towards,cisse2017parseval}; and 4) adversarial training, which adds adversarial samples into training procedure \cite{kurakin2016adversarial2,tramer2017ensemble,song2018improving} and can improve robustness of networks to a certain degree.
Goodfellow et al. \cite{goodfellow2014explaining} first increased the robustness of networks by feeding the model with both original and adversarial samples, and follow-up research modified it \cite{tramer2017ensemble,cai2018curriculum,kannan2018adversarial,wang2019bilateral,zhang2019defense}. 
Note that the adversarial learning with the form of GAN \cite{goodfellow2014generative} is not equivalent to adversarial training \cite{hung2018adversarial,luc2016semantic} and it cannot guarantee defense.

On the other hand, it is still rare in research to improve the robustness of semantic segmentation networks against various types of adversarial perturbations.
Xiao et al. \cite{xiao2018characterizing} proposed defense methods that aim at  detection of adversarial regions. We note detection only is not enough since the model still gives incorrect predictions. Several methods improve the robustness of segmentation networks with multitask learning \cite{klingner2020improved,mao2020multitask} and teacher-student frameworks \cite{bar2019robustness,bar2020robust}.
We advocate that models should accomplish correct output for adversarial samples during inference without extra training data and model parameters, and adversarial training is a universal method. But there is no comprehensive study of its effect on the semantic segmentation task.
Our proposed approach has stronger defense effect than state-of-the-art methods \cite{klingner2020improved,mao2020multitask,bar2019robustness,bar2020robust}.

\vspace{-0.05in}
\section{Standard Adversarial Attack}
\vspace{-0.05in}
Given a semantic segmentation network $f$ and an input $x$, the segmentation output is $o=f(x)$, where $x \in \mathbb{R}^{H \times W \times 3}$ and $o \in \mathbb{R}^{H \times W \times K}$ -- $H$, $W$ and $K$ are the height, width and number of classes respectively. 
For a clean sample $x^{clean}$, pixel $x^{clean}(i, j)$ is called ``clean pixel''; for the adversarial sample $x^{adv}$, which is obtained by adding perturbation on $x^{clean}$, pixel $x^{adv}(i, j)$ is called ``adversarial pixel'', paired with $x^{clean}(i, j)$.
The cross-entropy loss is denoted as $\mathcal{L}(f(x), y)$, where $y$ is the one-hot label of $x$.

Adversarial sample for $f$ can be generated by computing the gradient information of $f$ \cite{arnab2018robustness,goodfellow2014explaining}.
For example, given clean input $x^{clean}$ and its one-hot label $y$, FGSM attack \cite{goodfellow2014explaining} perturbs $x^{clean}$ as
\begin{equation}
\small
x^{adv}=x^{clean}+\epsilon \cdot sign(\bigtriangledown_{x^{clean}} (\mathcal{L}(f(x^{clean}), y))),
\end{equation}
where $x^{adv}$ is the resulting image with adversarial perturbation. $\epsilon$ constrains the level of perturbation. 
Further, iterative adversarial attack would cause more serious threat and BIM \cite{kurakin2016adversarial} is such an approach -- it has parameters for perturbation range $\epsilon$, step range $\alpha$, and start with $x^{adv_0} = x^{clean}$ -- as
\begin{equation}
\small
x^{adv_{t+1}}=\chi^{\epsilon}(x^{adv_t}+\alpha \cdot sign(\bigtriangledown_{x^{adv_t}} (\mathcal{L}(f(x^{adv_t}), y)))),
\end{equation} 
where $x^{adv_t}$ is the adversarial sample after the $t$-th attack step, function $\chi^{\epsilon}()$ forces its output to reside in the range of $[x^{clean}-\epsilon, x^{clean}+\epsilon]$, $sign()$ is the sign function and $\bigtriangledown_{a}(b)$ is the matrix derivative of $b$ with respect to $a$.

\begin{algorithm}[t]
	\caption{Standard Adversarial Training}
	\label{alg_baseline}
	\small
	{\bf Parameter:} 
	clean training set $\mathbf{X}$, segmentation network $f$, maximum number of training iterations $T_{max}$, $T \leftarrow 0$ 
	\begin{algorithmic}[1]
		\While{$T\not= T_{max}$}
		\parState {Load a mini-batch of data $\mathbf{D}_b = \{x^{clean}_1, ..., x^{clean}_m \}$ from the training set $\mathbf{X}$.}
		\parState {Get adversarial samples $\mathbf{A}_b = \{x^{adv}_1, ..., x^{adv}_m \}$ from $\mathbf{D}_b$.}
		\parState {Set batch as $\{x^{clean}_1, ..., x^{clean}_{\lfloor m/2 \rfloor}, x^{adv}_{\lfloor m/2 \rfloor+1}, ..., x^{adv}_m\}$ from $\mathbf{D}_b$ and $\mathbf{A}_b$, and compute the loss for this training batch. Update parameters of $f$. $T \leftarrow T + 1$.}
		\EndWhile
	\end{algorithmic}
\end{algorithm}

\begin{figure*}[t]
	\begin{center} 
		\includegraphics[width=0.9\linewidth]{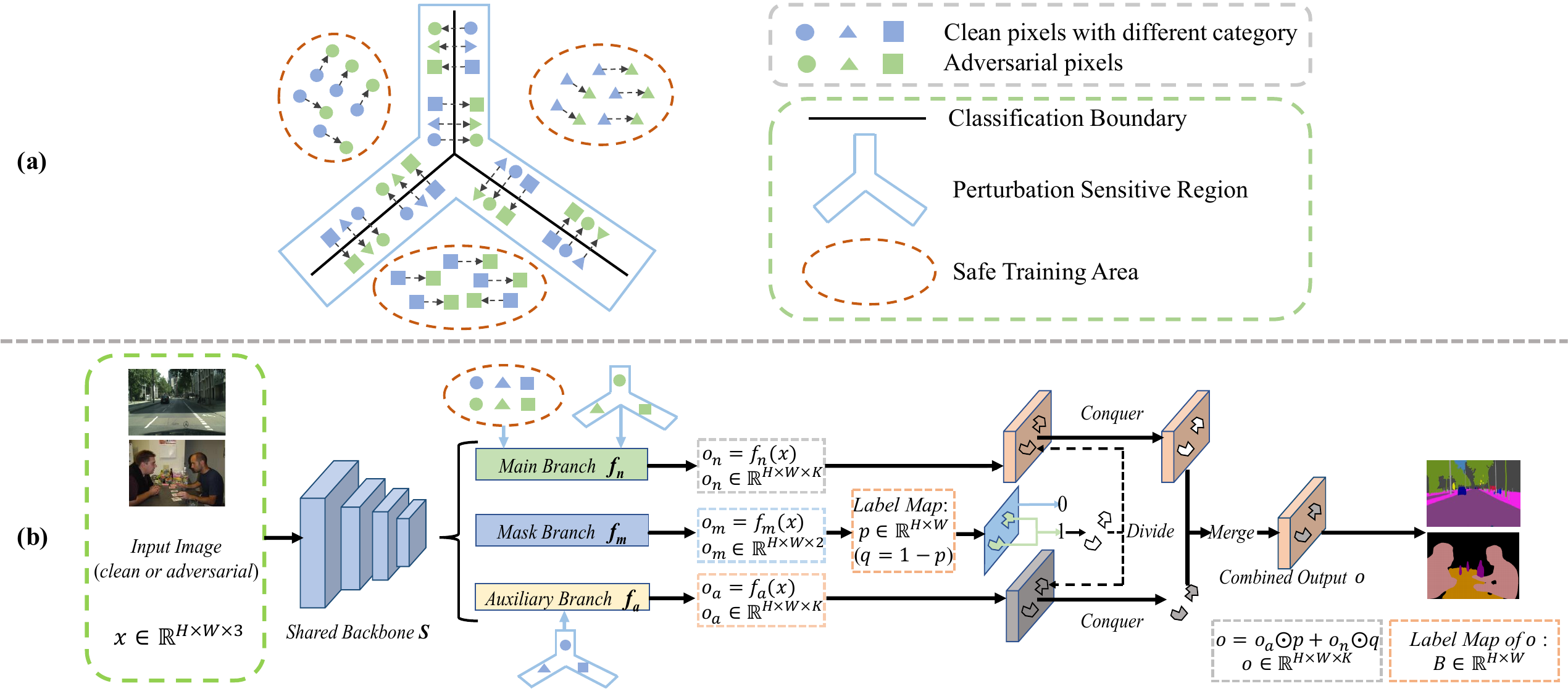}
	\end{center}
	\vspace{-0.1in}
	\caption{Motivation and the overall framework of DDC-AT (the arrow in (a) means perturbing clean pixels into adversarial pixels). (a) Clean pixels in the output space are divided into two categories by the divide-and-conquer strategy. (b) The main branch $f_n$ is utilized to conquer adversarial pixels and clean pixels stay far away from the classification boundary. The auxiliary branch $f_a$ is employed to conquer clean pixels that are sensitive to perturbation. The mask branch $f_m$ divides pixels into these two branches dynamically. The final output $o$ is combined from the division during training. During testing, both $f_a$ and $f_m$ are abandoned, and only $f_n$ is utilized to output $o_n$.}
	\label{fig:framework}
\end{figure*}

\vspace{-0.05in}
\section{Standard Adversarial Training}
\vspace{-0.05in}
We first design our standard adversarial training (SAT) on the semantic segmentation task. 
To ensure the performance on both clean/adversarial samples, we use mixed data where clean and adversarial samples are equally included in each batch during training.
This mixed strategy can scale up adversarial training to large models and datasets in classification \cite{kurakin2016adversarial2}. It also works for semantic segmentation.
The detailed procedure of SAT is listed in Alg. \ref{alg_baseline}. This algorithm yields reasonable defense effect on various datasets and meets part of our requirement.

\section{DDC-AT}
To further boost robustness of semantic segmentation networks, we propose a novel and much more effective strategy named dynamic divide-and-conquer adversarial training (DDC-AT).

\subsection{Divide-and-Conquer Procedure}
DDC-AT adopts a divide-and-conquer procedure during training, as shown in Fig. \ref{fig:framework} (b) and explained as the following.
1) Divide: for an input image $x$, DDC-AT divides its pixels into two sub-tasks for two branches, respectively. 2) Conquer: each branch predicts labels for the pixels assigned to it. 3) Merge: predictions from two branches are merged into the final prediction of image $x$.

\vspace{-0.1in}
\paragraph{Dividing pixels.} 
To improve the segmentation accuracy, previous work \cite{li2017not} has shown that we can divide pixels of one image into different kinds for individual processing, according to their property.
As shown in Fig. \ref{fig:framework}(a), clean pixels in the output space can be divided into two types during training, according to their ``boundary property''. 

1) Pixels $\mathcal{A}$ without ``boundary property'': clean pixels and their paired adversarial pixels are in the same classification space (in the ``Safe Training Area"). The properties of clean and adversarial pixels are similar in the output space. They are likely to stay far away from the boundary. 
Their distribution can be aligned in the same branch with adversarial training. 
The location of $\mathcal{A}$ in $x^{clean}$ is the set {\small $\{(i,j)| {\rm argmax}(f(x^{clean})(i,j))={\rm argmax}(f(x^{adv})(i,j)) \}$}.

2) Pixels $\mathcal{B}$ with ``boundary property'': clean pixels and their paired adversarial pixels are in diverse classification spaces. Such clean pixels are likely to stay near the classification boundary (in the ``Perturbation Sensitive Region"). They have ``boundary property'' since they are easy to perturb through the boundary. 
Directly aligning them with the adversarial pixels in the same branch is difficult, since their distributions differ widely.
We propose to first use two different branches to train them respectively. Once the clean and their adversarial pixels stay in the same space, we align them in the same branch.
Contrary to $\mathcal{A}$, the location of $\mathcal{B}$ in $x^{clean}$ is the set {\small $\{(i,j)| {\rm argmax}(f(x^{clean})(i,j))\neq {\rm argmax}(f(x^{adv})(i,j)) \}$}.

In short, we divide pixels in one clean image according to whether they have ``boundary property'' or not. The ``boundary property'' describes whether clean pixels and the corresponding adversarial pixels have different predictions or not.
For semantic segmentation, normally not all pixels in a clean sample are perturbed to have wrong predictions after adding adversarial noise \cite{xie2017adversarial,metzen2017universal,arnab2018robustness}. Thus, some pixels in a clean sample have the boundary property while others do not.
Such division can be completed dynamically via training a ``mask branch" $f_m$ that distinguishes among pixels with and without boundary property. Implementation of training $f_m$ will be discussed in Sec. \ref{sec:divide}. 

\vspace{-0.1in}
\paragraph{Conquering pixels.}
Based on above division setting, we set our framework as shown in Fig. \ref{fig:framework}(b), which consists of three branches. They are ``main branch'', ``auxiliary branch'', and ``mask branch'', denoted as $f_n$, $f_a$ and $f_m$ respectively. 
$f_n$ and $f_a$ can be utilized to \textit{conquer} pixels, i.e., predicting labels for pixels assigned to them through forwarding the corresponding networks.
We use ``main branch'' to conquer $\mathcal{A}$, as well as all adversarial pixels, and use ``auxiliary branch'' to conquer $\mathcal{B}$.
In this way, clean pixels in one image after division are processed by different branches. In addition, $f_n$ and $f_a$ share the backbone. Thus they help each other in the feature level. It is noteworthy that only $f_n$ is used in inference.

\vspace{-0.1in}
\paragraph{Merging pixels.}
As shown in Fig. \ref{fig:framework}(b), divided pixels after conquering are merged. 
This is because all pixels in one clean image are divided into $f_n$ and $f_a$, and there is no overlap between the pixels assigned to $f_n$ and $f_a$. Therefore, they can be merged into the final prediction of the input image to compute loss, according to the division.
This also indicates that the output space to decide the division should be the combination of $f_n$ and $f_a$ during training.

\begin{algorithm}[t]
	\caption{Algorithm to obtain ground truth (mask label) for training of mask branch $f_m$} 
	\label{alg1}
	\small
	\textbf{Parameter}: clean data $x^{clean}$ with one-hot label $y$, all-zero matrix $\textbf{0}$, function $\mathcal{F}=\textbf{1}[\mathcal{N}]$ ($\mathcal{F}(i, j)=1$ if $\mathcal{N}(i, j)$ is True)
	\begin{algorithmic}[1]
		\parState{\textbf{Obtain} output $o_n^{clean}$, $o_a^{clean}$, and $o_m^{clean}$ for $x^{clean}$ from $f_n$, $f_a$, and $f_m$. Label map of $o_m^{clean}$ is $p^{clean}$.}
		\parState{\textbf{Compute} $o^{clean} = o_a^{clean} \odot p^{clean} + o_n^{clean} \odot (1-p^{clean})$, its label map is $B^{clean}$, $B^{clean}(i, j) \in \{0, 1, ... K-1 \}$.}
		\parState{\textbf{Use} loss $\mathcal{L}(o_n^{clean}, y)$ to generate adversarial examples $x^{adv}$.}
		\parState{\textbf{Obtain} output $o_n^{adv}$, $o_a^{adv}$, and $o_m^{adv}$ for $x^{adv}$ from $f_n$, $f_a$, and $f_m$. The label map of $o_m^{adv}$ is $p^{adv}$.}
		\parState{\textbf{Compute} $o^{adv} = o_a^{adv} \odot p^{adv} + o_n^{adv} \odot (1-p^{adv})$ with label map $B^{adv}$, where $B^{adv}(i, j) \in \{0, 1, ... K-1 \}$.}
		\parState{\textbf{Generate} $M^{clean} = \textbf{1}[B^{clean} \neq B^{adv}]$, $M^{clean} \in \mathbb{R}^{H \times W}$.}
		\parState{\textbf{Generate} $M^{adv} = \textbf{0}$ with the same shape of $M^{clean}$.}
		\parState{\textbf{return} $M^{clean}$, $M^{adv}$, $x^{clean}$, and $x^{adv}$.}
	\end{algorithmic}
\end{algorithm}

\begin{algorithm*}[t]
	\caption{Dynamic divide-and-conquer adversarial training for semantic segmentation networks} 
	\label{alg2}
	\small
	\textbf{Parameter}: clean training set $\mathbf{X}$, shared backbone $S$, main branch $f_n$, auxiliary branch $f_a$, mask branch $f_m$, training batch size $m$, and maximum training iteration $T_{max}$, the number of iterations $T \leftarrow 0$
	\begin{algorithmic}[1]
		\While{$T\not= T_{max}$}
		\parState {\textbf{Load} a mini-batch of data $\mathbf{D}_b = \{x^{clean}_1, ..., x^{clean}_b \}$ from $\mathbf{X}$ with one-hot labels $\mathbf{Y}_b= \{y_1, ..., y_b \}$.}
		\parState {\textbf{Use} the current state of network $\{S, f_n, f_a, f_m \}$, $\mathbf{D}_b$, and $\mathbf{Y}_b$ to generate adversarial examples as $\mathbf{A}_b = \{x^{adv}_1, ..., x^{adv}_b \}$, and obtain ``mask label'' for $\mathbf{D}_b$ and $\mathbf{A}_b$ as $\mathbf{M}_b^{clean}=\{M^{clean}_1, ..., M^{clean}_b \}$ and $\mathbf{M}_b^{adv}=\{M^{adv}_1, ..., M^{adv}_b \}$.} 
		\parState{\textbf{Compute} output from $f_m$ for $\mathbf{D}_b$, and obtain the label map $\{p_1^{clean}, ..., p_b^{clean} \}$.}
		\parState{\textbf{Compute} output from $f_m$ for $\mathbf{A}_b$, and obtain the label map $\{p_1^{adv}, ..., p_b^{adv} \}$.}
		\parState{\textbf{Compute} $\{q_1^{clean}, ..., q_b^{clean}\}$ and $\{q_1^{adv}, ..., q_b^{adv}\}$, where $q_i^{clean}$ = $1-p_i^{clean}$, $q_i^{adv}$ = $1-p_i^{adv}$.}
		\parState {$\mathbf{T}_b=\{x^{clean}_1, ..., x^{clean}_{\lfloor b/2 \rfloor}, x^{adv}_{\lfloor b/2 \rfloor+1}, ..., x^{adv}_b\}$, 
			$\mathbf{M}_b=\{M^{clean}_1, ..., M^{clean}_{\lfloor b/2 \rfloor}, M^{adv}_{\lfloor b/2 \rfloor+1}, ..., M^{adv}_b\}$,\\ $\mathbf{P}_b=\{p_1^{clean}, ...,p_{\lfloor b/2 \rfloor}^{clean},p_{\lfloor b/2 \rfloor+1}^{adv}, ..., p_{b}^{adv}\}$, 
			$\mathbf{Q}_b=\{q_{1}^{clean}, ...,q_{\lfloor b/2 \rfloor}^{clean},q_{\lfloor b/2 \rfloor+1}^{adv}, ..., q_{b}^{adv}\}$.}
		\parState {\textbf{Compute} loss by \eqref{eq1_1} with $\mathbf{T}_b$, $\mathbf{Y}_b$, $\mathbf{P}_b$ and $\mathbf{Q}_b$. Update weights of network $\{S, f_n, f_a\}$.}
		\parState {\textbf{Compute} loss by \eqref{eq2} using $\mathbf{T}_b$ and $\mathbf{M}_b$.  Update weights of $\{S, f_m\}$. $T \leftarrow T + 1$.}
		\EndWhile
	\end{algorithmic}
\end{algorithm*}

\subsection{Dynamical Division and Implementation}
\label{sec:divide}
In this section, we illustrate the dynamical property of division setting in DDC-AT, and explain how such division is achieved through training a ``mask branch".

\vspace{-0.1in}
\paragraph{Dynamical division.}
Clean pixels belonging to $\mathcal{B}$ are first used in the auxiliary branch $f_a$ for training. 
They become robust towards adversarial perturbations in $f_a$ and turn into $\mathcal{A}$. Then they move to the main branch $f_n$. 
In this way, all clean pixels gradually move into $f_n$. 

In this design, the main branch finally trains all clean pixels, far away from the boundary. 
This mechanism effectively avoids drop of performance on clean samples. Moreover, training adversarial pixels with $\mathcal{A}$ improves robustness towards adversarial perturbation for the main branch.

\vspace{-0.1in}
\paragraph{Implementation.}
DDC-AT distributes all adversarial pixels into $f_n$, and adopts dynamical division for clean pixels. Such division is implemented with a ``mask branch'' $f_m$. 

\noindent\textbf{(a) Predicting $\mathcal{B}$.} First, $f_m$ predicts pixels $\mathcal{B}$ in the input image, as shown in Fig. \ref{fig:framework}(b).
For an input $x$, output from $f_n$, $f_a$, and $f_m$ is $o_n$, $o_a$, and $o_m \in \mathbb{R}^{H \times W \times 2}$ respectively. 
The label map of $o_m$ is $p \in \mathbb{R}^{H \times W}$, which is a binary matrix to decide division. $p(i,j)=1$ means pixel $x(i,j)$ belongs to $\mathcal{B}$ and is sent to $f_a$. Otherwise, it moves to $f_n$.
This operation yields the combined output for $x$ as $o=o_a \odot p + o_n \odot (1-p)$, as shown in Fig. \ref{fig:framework}(b). Here $\odot$ is the Hadamard product. If $x^{clean}$ is perturbed to $x^{adv}$, we denote the combined output as $o^{clean}$ and $o^{adv}$, which are obtained in the same way. 

\noindent\textbf{(b) Ground truth.} Next, the ideal division scheme is based on the combined output. This scheme has a ``mask label'' notation $M\in \mathbb{R}^{H \times W}$. $M(i, j)=1$ means the pixel in $(i, j)$ location belongs to $\mathcal{B}$ and is ``divided into $f_a$". Otherwise, it is ``divided into $f_n$". 
We set the mask label for $x^{clean}$ as $M^{clean}$, and denote the label map of $o^{clean}$ and $o^{adv}$ as $B^{clean}$ and $B^{adv}$ respectively.
For pixel $x^{clean}(i, j)$, if $B^{clean}(i,j)\neq B^{adv}(i,j)$, it sets into $f_a$ since it has the boundary property. In this case, we set $M^{clean}(i,j)=1$. Otherwise, it sends to $f_n$, making $M^{clean}(i,j)=0$.
Besides, all adversarial pixels should be sent to $f_n$, and we set mask label for $x^{adv}$ as $M^{adv} = \textbf{0}$, which is the matrix with all elements being zero.

\noindent\textbf{(c) Training.} 
$M^{adv}$ and $M^{clean}$ are obtained according to the ideal division rule in DDC-AT. We use them as the ground truth to train $f_m$. Repeating the whole process makes $f_m$ learn how to achieve ideal division for pixels automatically. 
The algorithm to obtain $M^{adv}$ and $M^{clean}$ is summarized in Alg. \ref{alg1}.
Note that learning of $f_m$ does not need external supervised information.

\noindent\textbf{(d) Results.} 
$\mathcal{B}$ turns to $\mathcal{A}$ and is assigned into $f_n$ progressively during training. This process is explained in Sec. \ref{sec:fm}.
Finally, almost all pixels are assigned into $f_n$, the decision boundary of $f_n$ tightly approaches that of the overall framework, and the predicted mask has almost all zero values.

\subsection{Overall Loss Function}
For the training data $x$ ($x^{clean}$ or $x^{adv}$), its label map obtained from the mask branch is $p \in \mathbb{R}^{H \times W}$, and we set $q = 1-p$.
The loss of $x$ for $f_n$ and $f_a$ is written as
\begin{equation}
\small
\begin{split}
&\mathcal{L}_{n} = \mathbb{E}\left( -\sum_{i=0}^{K-1} [y(:, :, i) \cdot \log(f_n(x)(:, :, i))]\odot q\right),\\
&\mathcal{L}_{a} = \mathbb{E}\left( -\sum_{i=0}^{K-1} [y(:, :, i) \cdot \log(f_a(x)(:, :, i))] \odot p\right),
\end{split}
\label{eq1_1}
\end{equation}
where $\mathbb{E}$ is the operation to compute the mean value, $\log()$ is the function of computing logarithm, $y(:, :, i)$, $f_n(x)(:, :, i)$ and $f_a(x)(:, :, i)$ are score maps with shape $\mathbb{R}^{H \times W}$.
Turning the mask label $M$ for $x$ into one-hot form $ \ \widetilde{M} \in \mathbb{R}^{H \times W \times 2}$, the loss for $f_m$ becomes
\begin{equation}
\small
\mathcal{L}_{m}= \mathbb{E}\left( -\sum_{i=0}^{1} [\widetilde{M}(:, :,i) \cdot \log(f_m(x)(:, :,i))]\right).
\label{eq2}
\end{equation} 
Combined with Eqs. \eqref{eq1_1} and \eqref{eq2}, the overall loss term is
\begin{equation}
\small
\mathcal{L}_{all} = \lambda_1 \mathcal{L}_{n} + \lambda_2 \mathcal{L}_{a} + \lambda_3 \mathcal{L}_{m},
\label{eq_final}
\end{equation}
where $\lambda_1$, $\lambda_2$, and $\lambda_3$ are set to 1 in experiments. Overall training procedure is concluded in Alg. \ref{alg2}.

\subsection{Superiority of Divide-and-Conquer}
DDC-AT is superior to SAT, proved in our experiments. 
It yields much better performance than classical SAT on both clean and adversarial pixels. We explain it below.

1) For segmentation, usually not all pixels in a clean sample are perturbed to have wrong predictions after adding adversarial noise. Some pixels have boundary property while others do not, and mask branch $f_m$ divides pixels into $f_n$ and $f_a$ based on their boundary properties. This motivates our division strategy.

2) First, for the training of clean pixels in the main branch $f_n$, training with $\mathcal{A}$ only (setting of DDC-AT) is much easier than mixed training with both $\mathcal{A}$ and $\mathcal{B}$ (setting of SAT). 
The introduced auxiliary branch $f_a$ in DDC-AT can turn $\mathcal{B}$ into $\mathcal{A}$ gradually and effectively. Thus, the main branch $f_n$ that is adopted for inference can better handle clean pixels and improve accuracy over SAT.

3) Second, to obtain decent results on adversarial pixels, SAT trains adversarial pixels with both $\mathcal{A}$ and $\mathcal{B}$, while DDC-AT trains adversarial pixels with only $\mathcal{A}$ for the main branch $f_n$. Obviously, training with both $\mathcal{A}$ and $\mathcal{B}$ causes higher difficulty for the learning of adversarial pixels than training with $\mathcal{A}$ only. Thus, DDC-AT yields higher accuracy on adversarial pixels.

\section{Experiments}
\label{sec:exp}
The newly proposed SAT and DDC-AT are effective for robust semantic segmentation.  We evaluate our method on challenging PASCAL VOC 2012 ~\cite{everingham2010pascal} and Cityscapes~\cite{cordts2016cityscapes} datasets, with popular semantic segmentation architectures PSPNet~\cite{zhao2017pyramid} and DeepLabv3~\cite{chen2017rethinking}. In the following, we first show implementation details related to training strategy and hyper-parameters. Then we exhibit results on corresponding datasets. The baseline is the method with no defense.

\subsection{Datasets}
PASCAL VOC 2012 (with abbreviation as VOC) \cite{everingham2010pascal} focuses on object segmentation. It contains 20 object classes and one class for background, with 1,464, 1,499, and 1,456 images for training, validation, and testing, respectively. The training set is augmented to 10,582 images in \cite{hariharan2015hypercolumns}, which is also adopted.
The Cityscapes \cite{cordts2016cityscapes} dataset is collected for urban scene understanding with 19 categories. It contains high-quality pixel-level annotations with 2,975, 500, and 1,525 images for training, validation, and testing.

\subsection{Implementation Details}
\label{sec:evaluation}
We choose popular semantic segmentation architectures PSPNet~\cite{zhao2017pyramid} and DeepLabv3~\cite{chen2017rethinking} for experiments. We follow the hyper-parameters as suggested in \cite{semseg2019} for all models. Both SAT and DDC-AT train networks with augmentation of adversarial samples. 

We choose white-box BIM attacker (by $L_{\infty}$ constraint) \cite{kurakin2016adversarial} to generate adversarial samples during training. The maximum perturbation value is set to $\epsilon=0.03\times 255$. The consideration is that perturbation can be visually noticed by human \cite{arnab2018robustness} with larger values.  The attack step size and number of attack iterations are set as $\alpha=0.01\times 255$ and $n=3$ for training respectively. 
We use the \textit{mean of class-wise intersection over union} (mIoU) as our evaluation metric. The parameters $\epsilon$, $\alpha$ and $n$ are kept constant during training. For each training mini-batch, half of the input includes adversarial samples that are dynamically decided by current model states, resulting in variance of results. For both SAT and DDC-AT, we train for one more time and report the average results as well as their standard deviations.

\begin{figure}[t]
	\centering
	\newcommand\widthpose{0.20}
	\newcommand\heightpose{0.16}
	\LARGE
	\resizebox{1.0\linewidth}{!}{
		\begin{tabular}{c@{\hspace{1.2mm}}c@{\hspace{1.2mm}}c@{\hspace{1.2mm}}c@{\hspace{1.2mm}}c@{\hspace{1.2mm}}c}
			
			\rotatebox{90}{\parbox[t]{15mm}{\hspace*{\fill} \hspace*{\fill}}}&
			\includegraphics[width=\widthpose\textwidth, height=\heightpose\textwidth]{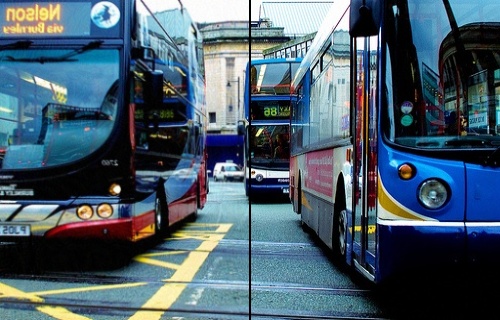} &
			\includegraphics[width=\widthpose\textwidth, height=\heightpose\textwidth]{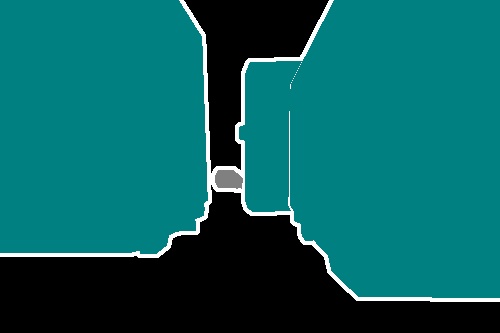} &
			\includegraphics[width=\widthpose\textwidth, height=\heightpose\textwidth]{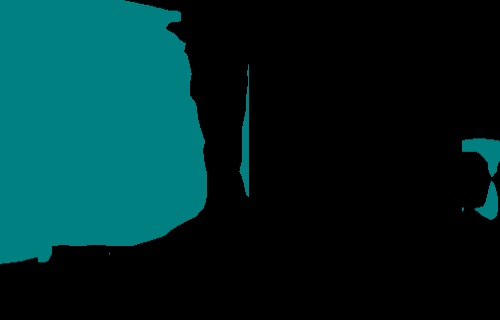} &
			\includegraphics[width=\widthpose\textwidth, height=\heightpose\textwidth]{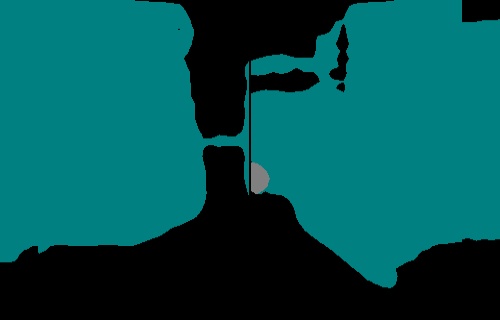} &
			\includegraphics[width=\widthpose\textwidth, height=\heightpose\textwidth]{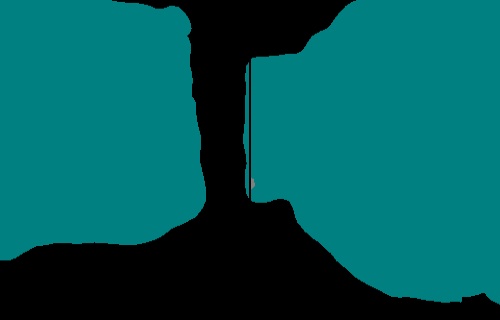} \\
			\rotatebox{90}{\parbox[t]{15mm}{\hspace*{\fill} \hspace*{\fill}}}&
			\includegraphics[width=\widthpose\textwidth, height=\heightpose\textwidth]{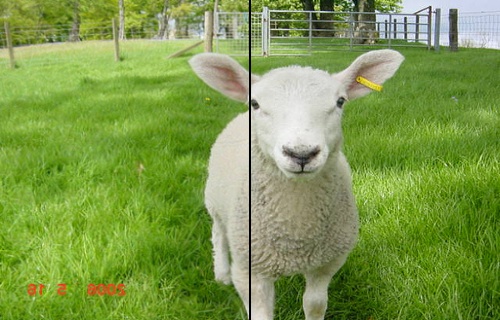} &
			\includegraphics[width=\widthpose\textwidth, height=\heightpose\textwidth]{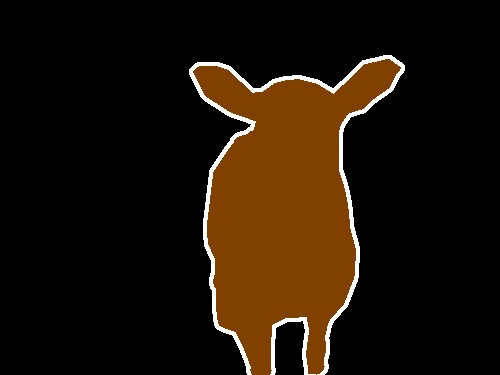} &
			\includegraphics[width=\widthpose\textwidth, height=\heightpose\textwidth]{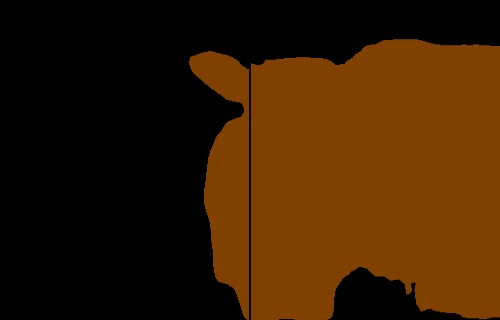} &
			\includegraphics[width=\widthpose\textwidth, height=\heightpose\textwidth]{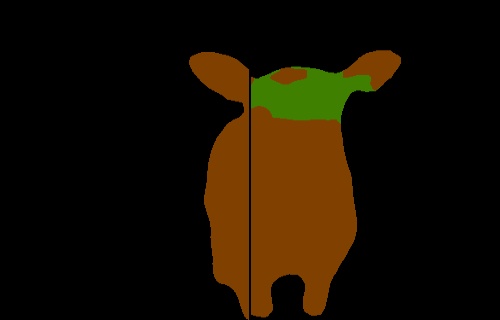} &
			\includegraphics[width=\widthpose\textwidth, height=\heightpose\textwidth]{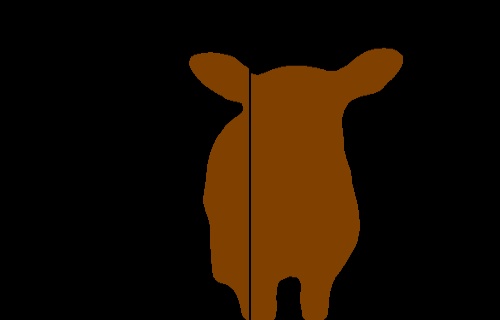} \\
			&{Image} &{Ground Truth} & {No Defense} & {SAT} &{DDC-AT} \\
	\end{tabular}}
	\caption{Visual comparison on VOC. The top row is obtained from models with PSPNet, and the bottom row is derived from models with DeepLabv3.}
	\label{fig:visual_voc}
\end{figure}

\subsection{PASCAL VOC 2012}
\vspace{-0.05in}
\paragraph{White-box attack.}
White-box attackers utilize the exact gradient information of the target model \cite{athalye2018robustness}. Specifically, for the evaluation, we consider untargeted BIM attack ($L_{\infty}$ constraint, $\epsilon=0.03\times 255$, $\alpha=0.01\times 255$) with $n$ ranging from 1 to 7, untargeted DeepFool attack ($L_{\infty}$ constraint, $\epsilon=0.03\times 255$) \cite{moosavi2016deepfool}, untargeted C\&W attack ($L_{\infty}$ constraint, $\epsilon=0.03\times 255$) \cite{carlini2017towards}, and untargeted BIM attack ($L_2$ constraint, $n$=3, $\epsilon=0.03\times 255$, $\alpha=0.01\times 255$).

\begin{table}[t]
	\caption{Evaluation under white-box attack on VOC. We report the mean value (Mean) and the standard deviation value (Std).
		``No Defense'' means normal training without adversarial samples. ``clean'' means mIoU (\%) on clean samples. Results in columns ``2'', ``4'', ``6'' are mIoUs (\%) under the BIM attack ($L_{\infty}$ constraint) with attack iteration number 2, 4, 6, respectively. Results in the column ``DeepFool", ``C\&W", ``BIM $L_2$" are mIoUs (\%) under DeepFool attack, C\&W attack, BIM attack ($L_2$ constraint).} 
	\label{tab:white-box-voc}
	\resizebox{1.0\linewidth}{!}{
		\begin{tabular}{c|c|cccccc}
			\toprule[1pt]
			&\multirow{2}{*}{{clean}} & \multicolumn{6}{c}{Model: PSPNet} \\
			\cline{3-8}
			& & 2 & 4 & 6 &{DeepFool}&{C\&W}&{BIM $L_2$}\\
			\hline
			No Defense &76.9&18.9 &7.8 &5.4&40.3&3.3&15.7\\
			SAT (Mean) &74.3 &68.1 &44.5 &27.9 &59.0&65.5&36.4 \\
			DDC-AT (Mean) &76.0 &\textbf{75.6} &\textbf{47.9} &\textbf{33.8} &\textbf{61.2}&\textbf{67.4}&\textbf{37.1}\\
			\hline
			SAT (Std) &0.5 &1.8 &2.9 &3.2 &1.4 &1.2&4.1 \\
			DDC-AT (Std) &0.1 &0.5 &2.2 &4.0 &1.1&1.1&1.8\\
			\hline
			&\multirow{2}{*}{{clean}} & \multicolumn{6}{c}{Model: DeepLabv3}\\
			\cline{3-8}
			& & 2 & 4 & 6  &{DeepFool}&{C\&W}&{BIM $L_2$}\\
			\hline
			No Defense &77.5&19.6 &8.1 &5.5  &39.3&3.9&16.7 \\
			SAT (Mean)&72.7 &62.4 &43.1 &28.8 &59.0&66.0&37.0\\
			DDC-AT(Mean)&75.2&\textbf{69.9} &\textbf{43.6} &\textbf{32.3} &\textbf{60.4}&\textbf{67.1}&\textbf{37.8}\\
			\hline
			SAT (Std)&1.0 &0.6 &1.9 &2.0 &0.4&1.2&1.1 \\
			DDC-AT (Std)&0.1 &1.3 &0.5 &1.2 &0.4&0.4&0.1\\
			\bottomrule[1pt]
	\end{tabular}}
\end{table}

Results of different defense methods on VOC are shown in Table \ref{tab:white-box-voc}. We compare our methods with the baseline (model trained with clean samples only, without defense). All untargeted attacks yield a sharp performance drop without defense. 
Especially, under untargeted BIM attack ($L_{\infty}$ constraint), the results approach zeros when the attack iteration number is large.

For BIM attack ($L_{\infty}$ constraint), Table \ref{tab:white-box-voc} basically indicates that results on adversarial samples with large attack iteration numbers represent the lower bound of each method on adversarial perturbation, since the corresponding performance decreases with the increase of $n$, and converges when $n$ is large. Actually, the mean value of mIoU does not change more than 2.5\% when $n$ is 10 or 20, compared with the results when $n$=6 for {\it No Defense}, SAT and DDC-AT.
This leads to the conclusion that SAT is already reasonable: it improves results from 5.4\% to 27.9\% on PSPNet and 5.5\% to 28.8\% on DeepLabv3 when $n=6$. 
Moreover, SAT also improves results on other different types of attacks.
Further, the standard deviation of SAT is low.

DDC-AT in Table \ref{tab:white-box-voc} gives results of our final framework. 
Performance of DDC-AT on clean samples increases compared with SAT (by 1.7\% and 2.5\% on PSPNet and DeepLabv3, respectively), consistent with our design motivation.
Further, the performance of DDC-AT is higher than SAT notably under each attacker iteration on average for BIM attack ($L_{\infty}$ constraint).
Intriguingly, the best case of SAT under every attack iteration is almost the worst case of DDC-AT. For example, when the attack iteration $n=2$, we have $68.1+1.8 < 75.6-0.5$ on PSPNet and $62.4+0.6 < 69.9-1.3$ on DeepLabv3. More interestingly, for unseen attacks, DDC-AT also clearly improves robustness over SAT. 
The small standard deviation for DDC-AT indicates that results are stable. We also provide the visual comparison on VOC in Fig. \ref{fig:visual_voc} for comparing result quality.

\begin{table}[t]
	\caption{Evaluation under black-box attack on VOC. Symbolic representations are the same as those in Table \ref{tab:white-box-voc}.}
	\label{tab:black-box-voc}
	\resizebox{1.0\linewidth}{!}{
		\begin{tabular}{c|c|cccccccccc}
			\toprule[1pt]
			&\multirow{2}{*}{{clean}} & \multicolumn{6}{c}{Model: PSPNet} \\
			\cline{3-8}
			& & 2 & 4 & 6 &{DeepFool}&{C\&W}&{BIM $L_2$}\\
			\hline
			No Defense &76.9 &24.0 &10.6 &6.0 &46.6&15.6&20.9\\
			SAT (Mean)     &74.3 &56.5 &51.3 &44.9 &64.0&68.5&58.7\\
			DDC-AT (Mean)&76.0&\textbf{61.5} &\textbf{53.4} &\textbf{46.1} &\textbf{68.4}&\textbf{70.5}&\textbf{59.6}\\
			\hline
			SAT (Std)&0.5 &2.9 &2.8 &4.2 &2.1&1.3&3.0 \\
			DDC-AT (Std)&0.1 &1.7 &1.8 &3.9 &0.3&0.1&0.8\\
			\hline
			&\multirow{2}{*}{{clean}} & \multicolumn{6}{c}{Model: DeepLabv3} \\
			\cline{3-8}
			& & 2 & 4 & 6 &{DeepFool}&{C\&W}&{BIM $L_2$}\\
			\hline
			No Defense &77.5 &24.6 &10.5 &7.0 &49.1&19.6&20.9 \\
			SAT (Mean)&72.7&51.8&51.0 &45.0 &64.4&68.5&64.5\\
			DDC-AT (Mean)&75.2 &\textbf{60.4} &\textbf{52.6} &\textbf{46.0} &\textbf{68.7}&\textbf{70.6}&\textbf{65.9}  \\
			\hline
			SAT (Std)&1.0 &3.8 &3.7 &4.1 &1.8&1.6&0.4 \\
			DDC-AT (Std)&0.1&5.1 &1.8 &1.6 &1.0&0.5&1.0  \\
			\bottomrule[1pt]
	\end{tabular}}
\end{table}

\vspace{-0.05in}
\paragraph{Black-box attack.}
Black-box attackers cannot utilize the exact gradient information of the target model. 
Instead, gradient information from a substitute network, which is defensively trained on the same dataset \cite{papernot2017practical,papernot2016transferability,liu2016delving}, can be adopted. 
In our evaluation setting, the perturbation for trained PSPNet models is generated by DeepLabv3, trained on the same dataset and enhanced with adversarial training, and vice versa. For SAT and DDC-AT, the substitute networks are the same.
As described in Sec. \ref{sec:evaluation}, models trained with the same method and dataset may demonstrate diverse behavior.

To reduce evaluation bias from training randomness, we evaluate SAT and DDC-AT on dataset $\widehat{D}$ in the following way. Using training strategy $\widehat{S}$ (SAT or DDC-AT) with a model structure $\widehat{f}$ on $\widehat{D}$, we obtain model set $\widehat{M}_1$. Then using adversarial training with a model structure different from $\widehat{f}$ on $\widehat{D}$, we obtain model set $\widehat{M}_2$ as substitute defensive networks. Finally, for each model in $\widehat{M}_1$, we use attack generated from each model in $\widehat{M}_2$ for evaluation.

The results under black-box evaluation on VOC are included in Table \ref{tab:black-box-voc}. The performance of clean models also decreases along with the increase of attack iteration for BIM attack ($L_{\infty}$ constraint), like the white-box situation. This phenomenon suggests that there is strong transferability for adversarial samples in the semantic segmentation task. Therefore, it is meaningful to evaluate robustness under this black-box setting.

In comparison between DDC-AT and SAT, we use the same hyper-parameters as white-box attacks. 
From Table \ref{tab:black-box-voc}, it is clear that SAT also improves the defense effect under black-box attacks. The standard deviation of SAT is larger than the results by white-box attacks because black-box perturbation for each model is obtained from a set of substitute networks, which yield different adversarial behaviors. 

The final performance of DDC-AT is consistently higher than SAT for BIM attack ($L_{\infty}$ constraint) as well as other attacks.
Meanwhile, standard deviations of DDC-AT are lower in all cases than SAT, especially under unseen attacks. It proves that DDC-AT is more stable than SAT by all types of attack. 

\begin{table}[t]
	\centering
	\caption{Performance comparison of defense setting in ablation study on VOC. Symbolic representation is the same as that of Table \ref{tab:white-box-voc}.} 
	\label{tab:abla}
	\resizebox{1\linewidth}{!}{
		\begin{tabular}{c|c|ccc|c|ccc}
			\toprule[1pt]
			& \multicolumn{4}{c|}{Model: PSPNet} &  \multicolumn{4}{c}{Model: DeepLabv3} \\
			\cline{2-9}
			&clean& 2 & 4 & 6 &clean & 2 & 4 & 6 \\
			\hline
			{SAT (Mean)}     &74.3&68.1 &44.5 &27.9  &72.7&62.4&43.1 &28.8 \\
			{DDC-AT-M (Mean)} &75.1 &69.0&44.1&30.4 &74.1 &\textbf{72.0} &\textbf{45.1} &31.4 \\
			{DDC-AT-N (Mean)} &74.8 &73.7&45.6&31.8& 74.1 &68.3 &42.6 &31.2\\
			{DDC-AT (Mean)} &76.0&\textbf{75.6} &\textbf{47.9} &\textbf{33.8} &75.2  &69.9  &43.6&\textbf{32.3}\\
			\hline
			{SAT (Std)}     &0.5 &1.8&2.9 &3.2 &1.0 &0.6 &1.9 &2.0  \\
			{DDC-AT-M (Std)} &0.1 &3.1&0.9&4.0 &0.1 &4.6 &4.7 &3.8 \\
			{DDC-AT-N (Std)} &0.3 &0.3&0.3&3.4& 0.1 &4.4 &2.4  &1.7 \\
			{DDC-AT (Std)} &0.1 &0.5&2.2 &4.0 &0.1 &1.3 &0.5 &1.2\\
			\bottomrule[1pt]
	\end{tabular}}
\end{table}

\subsection{Ablation Study}
The motivation of DDC-AT is to dynamically divide pixels with/without boundary property into diverse branches during training. We prove our division setting is better than other alternatives by adjusting the division setting for pixels with boundary property. The alternatives are the following.

1) Use the ``main branch'' to deal with pixels from clean and adversarial samples without boundary property. Use ``auxiliary branch'' to process pixels from clean and adversarial samples with boundary property. We name this setting as DDC-AT-M. 

2) Use the ``main branch'' to deal with pixels from clean samples, pixels from adversarial samples without boundary property. Use ``auxiliary branch'' to solve pixels from adversarial samples with boundary property. We name this setting as DDC-AT-N.

3)Use only the ``main branch'' to deal with pixels from either clean or adversarial samples. This is what SAT does, thus we do not train the mask branch.

For all these methods, only the main branch is utilized during testing. We evaluate the performances of these alternatives and list results in Table \ref{tab:abla}. For PSPNet model, the performance of DDC-AT-N is higher than SAT and lower than DDC-AT. Their standard deviations are in the same scale. The average results of DDC-AT-M are comparable with SAT and are worse than DDC-AT. Also, compared with DDC-AT, the standard deviation increases clearly by DDC-AT-M. This is because the adversarial samples during the training are different at every training iteration, and the dynamical distribution enhances such randomness. Similarly, for DeepLabv3 model, the average results of DDC-AT-M and DDC-AT-N are lower than DDC-AT, and higher than SAT consistently. The standard deviation increases compared with DDC-AT and SAT.
In summary, the division setting of DDC-AT is optimal among these alternatives in terms of average performance and stability measurement.

\begin{table}[t]
	\caption{Evaluation under white-box attack on Cityscapes. Symbolic representation is the same as that of Table \ref{tab:white-box-voc}.}
	\label{tab:white-box-city}
	\resizebox{1.0\linewidth}{!}{
		\begin{tabular}{c|c|cccccc}
			\toprule[1pt]
			&\multirow{2}{*}{{clean}}   & \multicolumn{6}{c}{Model: PSPNet} \\
			\cline{3-8}
			&& 2 & 4 & 6  &{DeepFool}&{C\&W}&{BIM $L_2$}\\
			\hline
			No Defense & 74.6&26.2 &5.5 &2.1 &35.8&13.8&22.7 \\
			SAT (Mean) &69.0 &46.7 &32.9 &25.8 &56.0&49.1&45.8 \\
			DDC-AT (Mean) &71.7 &\textbf{50.2} &\textbf{34.7} &\textbf{28.7} &\textbf{57.2}&\textbf{50.8}&\textbf{46.7}\\
			\hline
			SAT (Std) &1.0 &1.0 &0.3 &1.0 &3.0&1.5&1.8 \\
			DDC-AT (Std) &0.1  &0.2  &0.2 &0.3 &0.1&0.1&0.1\\
			\hline
			&\multirow{2}{*}{{clean}}  & \multicolumn{6}{c}{Model: DeepLabv3} \\
			\cline{3-8}
			&& 2 & 4 & 6 &{DeepFool}&{C\&W}&{BIM $L_2$}\\
			\hline
			No Defense  &74.8 &26.0 &5.7 &2.3 &31.5&13.8&22.6  \\
			SAT (Mean)&69.4 &46.1 &31.8 &26.2 &56.7&48.4&45.0 \\
			DDC-AT (Mean) &71.3 &\textbf{50.9} &\textbf{34.9}  &\textbf{29.0} &\textbf{57.4}&\textbf{50.5}&\textbf{46.8} \\
			\hline
			SAT (Std)&1.0 &1.0  &0.6 &0.4 &1.7&1.3&0.9\\
			DDC-AT (Std) &0.3 &0.4 &0.2&0.2 &0.2&1.5&0.4\\
			\bottomrule[1pt]
	\end{tabular}}
\end{table}

\begin{table}[t]
	\caption{Evaluation under black-box attack on Cityscapes.}
	\label{tab:black-box-city}
	\resizebox{1.0\linewidth}{!}{
		\begin{tabular}{c|c|cccccc}
			\toprule[1pt]
			&\multirow{2}{*}{{clean}}   & \multicolumn{6}{c}{Model: PSPNet} \\
			\cline{3-8}
			&& 2 & 4 & 6 &{DeepFool}&{C\&W}&{BIM $L_2$} \\
			\hline
			No Defense &74.6&28.0 &6.9 &3.3 &35.6&21.1&25.3 \\
			SAT (Mean) &69.0&44.4&36.7 &30.8 &57.7&57.8&56.6 \\
			DDC-AT (Mean) &71.7 &\textbf{50.6} &\textbf{37.9}&\textbf{32.3} &\textbf{58.6}&\textbf{58.4}&\textbf{57.4}\\
			\hline
			SAT (Std) &1.0&3.0 &3.3 &2.6  &2.4&2.0&2.5 \\
			DDC-AT (Std) &0.1 &1.0 &1.0 &0.2 &0.1&0.1&0.3\\
			\hline
			&\multirow{2}{*}{{clean}}  & \multicolumn{6}{c}{Model: DeepLabv3} \\
			\cline{3-8}
			&& 2 & 4 & 6&{DeepFool}&{C\&W}&{BIM $L_2$} \\
			\hline
			No Defense  &74.8 &29.9  &7.6 &3.1 &35.8&27.3&27.3\ \\
			SAT (Mean)&69.4&43.2&36.1 &31.6 &58.4&58.3&57.4\ \\
			DDC-AT (Mean) &71.3 &\textbf{47.8}  &\textbf{37.8} &\textbf{32.8}  &\textbf{59.6}& \textbf{59.7}& \textbf{59.2}\\
			\hline
			SAT (Std)&1.0 &3.0  &3.0 &2.3 &1.1&1.8&2.3 \\
			DDC-AT (Std)&0.3 &1.9 &1.0 &0.3 &0.7&0.5&0.2\\
			\bottomrule[1pt]
	\end{tabular}}
\end{table}

\begin{table}[t]
	\caption{Evaluation under white- and black-box attack on Cityscapes.}
	\label{tab:white-box-city2}
	\resizebox{1.0\linewidth}{!}{
		\begin{tabular}{c|cccc|cccc}
			\toprule[1pt]
			& \multicolumn{4}{c|}{PSPNet (white-box)}& \multicolumn{4}{c}{PSPNet (black-box)} \\
			\cline{2-9}
			& 2 &\scriptsize{DeepFool}&\scriptsize{C\&W}&\scriptsize{BIM $L_2$}& 2 &\scriptsize{DeepFool}&\scriptsize{C\&W}&\scriptsize{BIM $L_2$}\\
			\hline
			multi-task \cite{klingner2020improved} &38.4&40.6&26.3&34.2&40.1&42.4&28.6&35.5\\
			multi-task \cite{mao2020multitask} &30.3&37.6&17.3&25.8&31.4&38.2&23.6&27.3\\
			TS \cite{bar2019robustness} &41.6&54.3&40.4&43.8&43.2&55.7&42.6&49.2\\
			TS \cite{bar2020robust} &47.9&56.8&44.5&45.2&48.3&57.1&47.2&51.8\\
			DDC-AT &\textbf{50.2} &\textbf{57.2}&\textbf{50.8}&\textbf{46.7} &\textbf{50.6}  &\textbf{58.6}&\textbf{58.4}&\textbf{57.4}\\
			\hline
			& \multicolumn{4}{c|}{DeepLabv3 (white-box)}& \multicolumn{4}{c}{DeepLabv3 (black-box)}  \\
			\cline{2-9}
			&2 &\scriptsize{DeepFool}&\scriptsize{C\&W}&\scriptsize{BIM $L_2$}&2 &\scriptsize{DeepFool}&\scriptsize{C\&W}&\scriptsize{BIM $L_2$}\\
			\hline
			multi-task \cite{klingner2020improved} &37.5&41.2&27.4&35.8&41.6&44.1&32.3&37.9\\
			multi-task \cite{mao2020multitask} &28.9&35.3&16.8&25.5&31.8&38.7&30.2&31.4\\
			TS \cite{bar2019robustness} &42.3&54.0&41.1&42.4&44.5&56.2&44.8&51.5\\
			TS \cite{bar2020robust} &48.1&55.3&46.5&45.3&50.3&57.6&50.3&53.7\\
			DDC-AT&\textbf{50.9} &\textbf{57.4}&\textbf{50.5}&\textbf{46.8} &\textbf{47.8}  &\textbf{59.6}& \textbf{59.7}& \textbf{59.2}\\
			\bottomrule[1pt]
	\end{tabular}}
\end{table}

\subsection{Cityscapes}
\vspace{-0.05in}
\paragraph{White-box attack.}
The results of different methods on clean samples are included in Table \ref{tab:white-box-city}. DDC-AT effectively reduces the drop of performance on clean samples, compared with SAT. DDC-AT improves mIoU on clean samples by 2.7\% and 1.9\%, which are significant with the setting of PSPNet and DeepLabv3, compared with SAT. The results of DDC-AT are also more stable over SAT.

We show results under white-box attack on Cityscapes dataset in Table \ref{tab:white-box-city}.
Obviously, clean models get worse with the increase of attack iterations for BIM attack ($L_{\infty}$ constraint), like the case in VOC, which proves the general effect of the adversarial attack on different datasets.
The results of DDC-AT and SAT under various attack iterations for BIM attack ($L_{\infty}$ constraint) are like we observe before -- they also improve the robustness of the models on this large dataset. 

DDC-AT outperforms SAT in Table \ref{tab:white-box-city} where the best cases of SAT under every attack iteration are actually worse than the worst cases of DDC-AT. 
Further, DDC-AT also outperforms SAT on other types of attacks in this dataset. 
As shown in Table \ref{tab:white-box-city}, both DDC-AT and SAT are stable, while DDC-AT is even better.

Besides, some current methods \cite{klingner2020improved,mao2020multitask,bar2019robustness,bar2020robust} can also achieve defense effects on Cityscapes. The comparison with them is reported in Table \ref{tab:white-box-city2}, demonstrating the superiority of our approach.

\vspace{-0.05in}
\paragraph{Black-box attack.}
The results under the evaluation of the black-box attack for the Cityscapes dataset are shown in Table \ref{tab:black-box-city}. 
DDC-AT also outperforms SAT on average with various types of attacks. The standard deviation of DDC-AT is still strictly smaller than that of SAT. For example, when $n=6$, the standard deviation of SAT is larger than 2\% for both PSPNet and DeepLabv3, while the standard deviation of DDC-AT is smaller than 0.5\%.
Our method also outperforms \cite{klingner2020improved,mao2020multitask,bar2019robustness,bar2020robust} as shown in Table \ref{tab:white-box-city2}.

\begin{figure}[t]
	\includegraphics[width=1.0\linewidth]{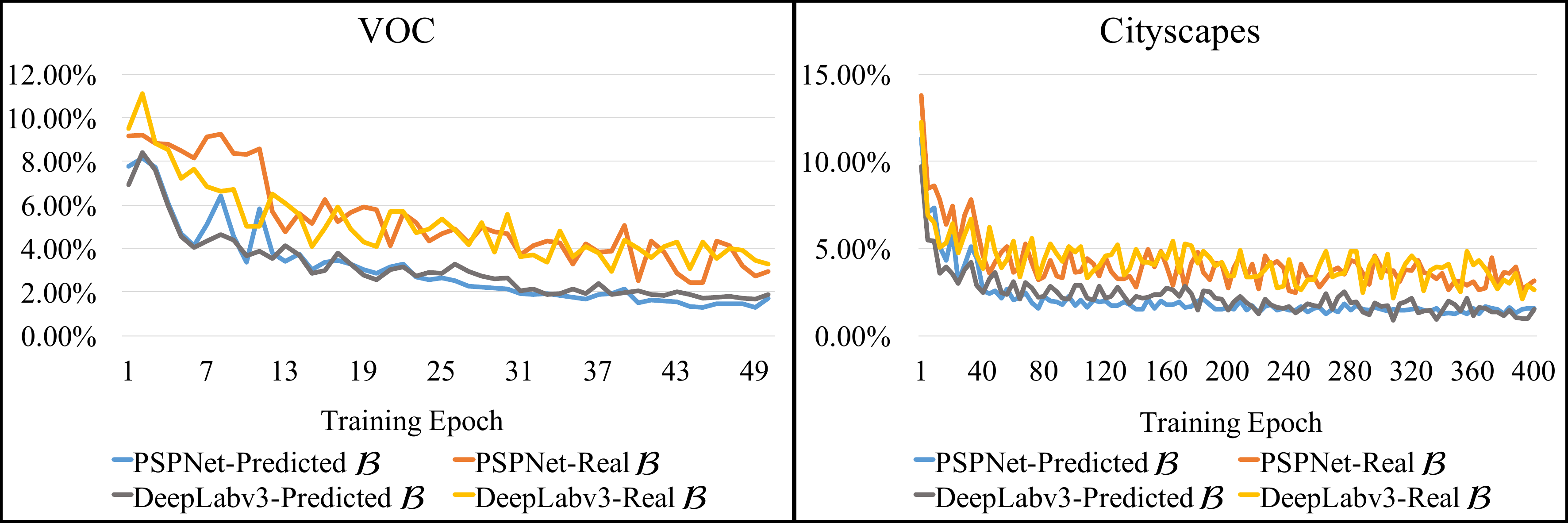}
	\vspace{-0.1in}
	\caption{The proportion of predicted/real $\mathcal{B}$ in one clean image with respect to the number of training epoch.}
	\label{fig:mask}
\end{figure}

\subsection{The Output of $f_m$}
\label{sec:fm}
As we have mentioned, clean pixels with the boundary property ($\mathcal{B}$) are first set to $f_a$ for training, and they will gradually be turned into clean pixels without the boundary property ($\mathcal{A}$).
The number of the predicted $\mathcal{B}$ in one clean image is decided by the output of $f_m$, i.e., the total number of non-zero values in the label map $p$, as clarified in Sec. \ref{sec:divide}.
And the number of the real $\mathcal{B}$ can be computed as the total number of non-zero values in the mask label $M$, as clarified in Sec. \ref{sec:divide} and Alg. \ref{alg1}.
We display the average proportion of the predicted/real $\mathcal{B}$ in one clean image with respect to the number of training epoch in Fig. \ref{fig:mask}.
Obviously, the proportion of the predicted and real $\mathcal{B}$ gradually approaches zeros. The tendency of the predicted $\mathcal{B}$ follows the real $\mathcal{B}$, since the mask branch $f_m$ is supervised by the mask label $M$, as shown in Eq. \eqref{eq2}.
All these results show that $\mathcal{B}$ gradually turns into $\mathcal{A}$.

\section{Conclusion}
In this paper, we have explored the property of adversarial training on the semantic segmentation task.
Our defense strategy can consistently enhance the robustness of target models under adversarial attacks. 
Besides proposing the standard adversarial training (SAT) process, we propose a new strategy to improve the performance of adversarial training in this task, with no extra parameter and computation cost introduced during inference.
The extensive experimental results with different model structures on two representative benchmark datasets suggest that the proposed method achieves significantly better generalization and stability on unseen adversarial examples and clean samples, compared with standard adversarial training. 

{\small
\bibliographystyle{ieee_fullname}
\bibliography{egbib}

\begin{thebibliography}{10}\itemsep=-1pt

\bibitem{arnab2018robustness}
Anurag Arnab, Ondrej Miksik, and Philip~HS Torr.
\newblock On the robustness of semantic segmentation models to adversarial
  attacks.
\newblock In {\em CVPR}, 2018.

\bibitem{athalye2018robustness}
Anish Athalye and Nicholas Carlini.
\newblock On the robustness of the cvpr 2018 white-box adversarial example
  defenses.
\newblock {\em arXiv:1804.03286}, 2018.

\bibitem{bar2019robustness}
Andreas Bar, Fabian Huger, Peter Schlicht, and Tim Fingscheidt.
\newblock On the robustness of redundant teacher-student frameworks for
  semantic segmentation.
\newblock In {\em CVPRW}, 2019.

\bibitem{bar2020robust}
Andreas Bar, Marvin Klingner, Serin Varghese, Fabian Huger, Peter Schlicht, and
  Tim Fingscheidt.
\newblock Robust semantic segmentation by redundant networks with a
  layer-specific loss contribution and majority vote.
\newblock In {\em CVPRW}, 2020.

\bibitem{cai2018curriculum}
Qi-Zhi Cai, Min Du, Chang Liu, and Dawn Song.
\newblock Curriculum adversarial training.
\newblock {\em IJCAI}, 2018.

\bibitem{carlini2017towards}
Nicholas Carlini and David Wagner.
\newblock Towards evaluating the robustness of neural networks.
\newblock In {\em IEEE symposium on security and privacy}, 2017.

\bibitem{chen2017rethinking}
Liang-Chieh Chen, George Papandreou, Florian Schroff, and Hartwig Adam.
\newblock Rethinking atrous convolution for semantic image segmentation.
\newblock {\em arXiv:1706.05587}, 2017.

\bibitem{cisse2017parseval}
Moustapha Cisse, Piotr Bojanowski, Edouard Grave, Yann Dauphin, and Nicolas
  Usunier.
\newblock Parseval networks: Improving robustness to adversarial examples.
\newblock In {\em ICML}, 2017.

\bibitem{cordts2016cityscapes}
Marius Cordts, Mohamed Omran, Sebastian Ramos, Timo Rehfeld, Markus Enzweiler,
  Rodrigo Benenson, Uwe Franke, Stefan Roth, and Bernt Schiele.
\newblock The cityscapes dataset for semantic urban scene understanding.
\newblock In {\em CVPR}, 2016.

\bibitem{dhillon2018stochastic}
Guneet~S Dhillon, Kamyar Azizzadenesheli, Zachary~C Lipton, Jeremy Bernstein,
  Jean Kossaifi, Aran Khanna, and Anima Anandkumar.
\newblock Stochastic activation pruning for robust adversarial defense.
\newblock In {\em ICLR}, 2018.

\bibitem{ding2019defending}
Yifan Ding, Liqiang Wang, Huan Zhang, Jinfeng Yi, Deliang Fan, and Boqing Gong.
\newblock Defending against adversarial attacks using random forest.
\newblock In {\em CVPRW}, 2019.

\bibitem{dong2018boosting}
Yinpeng Dong, Fangzhou Liao, Tianyu Pang, Hang Su, Jun Zhu, Xiaolin Hu, and
  Jianguo Li.
\newblock Boosting adversarial attacks with momentum.
\newblock In {\em CVPR}, 2018.

\bibitem{everingham2010pascal}
Mark Everingham, Luc Van~Gool, Christopher~KI Williams, John Winn, and Andrew
  Zisserman.
\newblock The pascal visual object classes (voc) challenge.
\newblock {\em IJCV}, 2010.

\bibitem{goodfellow2014generative}
Ian Goodfellow, Jean Pouget-Abadie, and Mehdi Mirza.
\newblock Generative adversarial nets.
\newblock In {\em NIPS}, 2014.

\bibitem{goodfellow2014explaining}
Ian~J Goodfellow, Jonathon Shlens, and Christian Szegedy.
\newblock Explaining and harnessing adversarial examples.
\newblock {\em ICLR}, 2014.

\bibitem{gu2014towards}
Shixiang Gu and Luca Rigazio.
\newblock Towards deep neural network architectures robust to adversarial
  examples.
\newblock {\em arXiv:1412.5068}, 2014.

\bibitem{guo2017countering}
Chuan Guo, Mayank Rana, Moustapha Cisse, and Laurens Van Der~Maaten.
\newblock Countering adversarial images using input transformations.
\newblock In {\em ICLR}, 2018.

\bibitem{hariharan2015hypercolumns}
Bharath Hariharan, Pablo Arbel{\'a}ez, Ross Girshick, and Jitendra Malik.
\newblock Hypercolumns for object segmentation and fine-grained localization.
\newblock In {\em CVPR}, 2015.

\bibitem{hung2018adversarial}
Wei-Chih Hung, Yi-Hsuan Tsai, and Yan-Ting Liou.
\newblock Adversarial learning for semi-supervised semantic segmentation.
\newblock In {\em BMVC}, 2018.

\bibitem{inkawhich2019feature}
Nathan Inkawhich, Wei Wen, Hai~Helen Li, and Yiran Chen.
\newblock Feature space perturbations yield more transferable adversarial
  examples.
\newblock In {\em CVPR}, 2019.

\bibitem{jia2019comdefend}
Xiaojun Jia, Xingxing Wei, Xiaochun Cao, and Hassan Foroosh.
\newblock Comdefend: An efficient image compression model to defend adversarial
  examples.
\newblock In {\em CVPR}, 2019.

\bibitem{kannan2018adversarial}
Harini Kannan, Alexey Kurakin, and Ian Goodfellow.
\newblock Adversarial logit pairing.
\newblock {\em arXiv:1803.06373}, 2018.

\bibitem{klingner2020improved}
Marvin Klingner, Andreas Bar, and Tim Fingscheidt.
\newblock Improved noise and attack robustness for semantic segmentation by
  using multi-task training with self-supervised depth estimation.
\newblock In {\em CVPRW}, 2020.

\bibitem{kurakin2016adversarial}
Alexey Kurakin, Ian Goodfellow, and Samy Bengio.
\newblock Adversarial examples in the physical world.
\newblock {\em ICLR}, 2016.

\bibitem{kurakin2016adversarial2}
Alexey Kurakin, Ian Goodfellow, and Samy Bengio.
\newblock Adversarial machine learning at scale.
\newblock {\em ICLR}, 2016.

\bibitem{li2017not}
Xiaoxiao Li, Ziwei Liu, Ping Luo, Chen Change~Loy, and Xiaoou Tang.
\newblock Not all pixels are equal: Difficulty-aware semantic segmentation via
  deep layer cascade.
\newblock In {\em CVPR}, 2017.

\bibitem{liu2016delving}
Yanpei Liu, Xinyun Chen, Chang Liu, and Dawn Song.
\newblock Delving into transferable adversarial examples and black-box attacks.
\newblock In {\em ICLR}, 2017.

\bibitem{luc2016semantic}
Pauline Luc, Camille Couprie, and Soumith Chintala.
\newblock Semantic segmentation using adversarial networks.
\newblock {\em arXiv:1611.08408}, 2016.

\bibitem{madry2017towards}
Aleksander Madry, Aleksandar Makelov, Ludwig Schmidt, Dimitris Tsipras, and
  Adrian Vladu.
\newblock Towards deep learning models resistant to adversarial attacks.
\newblock In {\em ICLR}, 2018.

\bibitem{mao2020multitask}
Chengzhi Mao, Amogh Gupta, Vikram Nitin, Baishakhi Ray, Shuran Song, Junfeng
  Yang, and Carl Vondrick.
\newblock Multitask learning strengthens adversarial robustness.
\newblock {\em ECCV}, 2020.

\bibitem{metzen2017universal}
Jan~Hendrik Metzen, Mummadi~Chaithanya Kumar, Thomas Brox, and Volker Fischer.
\newblock Universal adversarial perturbations against semantic image
  segmentation.
\newblock In {\em ICCV}, 2017.

\bibitem{moosavi2016deepfool}
Seyed-Mohsen Moosavi-Dezfooli, Alhussein Fawzi, and Pascal Frossard.
\newblock Deepfool: a simple and accurate method to fool deep neural networks.
\newblock In {\em CVPR}, 2016.

\bibitem{papernot2016transferability}
Nicolas Papernot, Patrick McDaniel, and Ian Goodfellow.
\newblock Transferability in machine learning: from phenomena to black-box
  attacks using adversarial samples.
\newblock {\em arXiv:1605.07277}, 2016.

\bibitem{papernot2017practical}
Nicolas Papernot, Patrick McDaniel, Ian Goodfellow, Somesh Jha, Z~Berkay Celik,
  and Ananthram Swami.
\newblock Practical black-box attacks against machine learning.
\newblock In {\em ACM on Asia conference on computer and communications
  security}, 2017.

\bibitem{papernot2016limitations}
Nicolas Papernot, Patrick McDaniel, Somesh Jha, Matt Fredrikson, Z~Berkay
  Celik, and Ananthram Swami.
\newblock The limitations of deep learning in adversarial settings.
\newblock In {\em 2016 IEEE European Symposium on Security and Privacy}, 2016.

\bibitem{prakash2018deflecting}
Aaditya Prakash, Nick Moran, Solomon Garber, Antonella DiLillo, and James
  Storer.
\newblock Deflecting adversarial attacks with pixel deflection.
\newblock In {\em CVPR}, 2018.

\bibitem{raff2019barrage}
Edward Raff, Jared Sylvester, Steven Forsyth, and Mark McLean.
\newblock Barrage of random transforms for adversarially robust defense.
\newblock In {\em CVPR}, 2019.

\bibitem{song2018improving}
Chuanbiao Song, Kun He, Liwei Wang, and John~E Hopcroft.
\newblock Improving the generalization of adversarial training with domain
  adaptation.
\newblock {\em ICLR}, 2018.

\bibitem{song2017pixeldefend}
Yang Song, Taesup Kim, Sebastian Nowozin, Stefano Ermon, and Nate Kushman.
\newblock Pixeldefend: Leveraging generative models to understand and defend
  against adversarial examples.
\newblock In {\em ICLR}, 2018.

\bibitem{szegedy2013intriguing}
Christian Szegedy, Wojciech Zaremba, Ilya Sutskever, Joan Bruna, Dumitru Erhan,
  Ian Goodfellow, and Rob Fergus.
\newblock Intriguing properties of neural networks.
\newblock {\em arXiv:1312.6199}, 2013.

\bibitem{tramer2017ensemble}
Florian Tram{\`e}r, Alexey Kurakin, Nicolas Papernot, Ian Goodfellow, Dan
  Boneh, and Patrick McDaniel.
\newblock Ensemble adversarial training: Attacks and defenses.
\newblock {\em ICLR}, 2017.

\bibitem{wang2019bilateral}
Jianyu Wang and Haichao Zhang.
\newblock Bilateral adversarial training: Towards fast training of more robust
  models against adversarial attacks.
\newblock In {\em ICCV}, 2019.

\bibitem{wu2020boosting}
Weibin Wu, Yuxin Su, Xixian Chen, Shenglin Zhao, Irwin King, Michael~R Lyu, and
  Yu-Wing Tai.
\newblock Boosting the transferability of adversarial samples via attention.
\newblock In {\em CVPR}, 2020.

\bibitem{xiao2018characterizing}
Chaowei Xiao, Ruizhi Deng, Bo Li, Fisher Yu, Mingyan Liu, and Dawn Song.
\newblock Characterizing adversarial examples based on spatial consistency
  information for semantic segmentation.
\newblock In {\em ECCV}, 2018.

\bibitem{xie2017mitigating}
Cihang Xie, Jianyu Wang, Zhishuai Zhang, Zhou Ren, and Alan Yuille.
\newblock Mitigating adversarial effects through randomization.
\newblock {\em ICLR}, 2017.

\bibitem{xie2017adversarial}
Cihang Xie, Jianyu Wang, Zhishuai Zhang, Yuyin Zhou, Lingxi Xie, and Alan
  Yuille.
\newblock Adversarial examples for semantic segmentation and object detection.
\newblock In {\em CVPR}, 2017.

\bibitem{xie2019feature}
Cihang Xie, Yuxin Wu, Laurens van~der Maaten, Alan~L Yuille, and Kaiming He.
\newblock Feature denoising for improving adversarial robustness.
\newblock In {\em CVPR}, 2019.

\bibitem{zhang2019defense}
Haichao Zhang and Jianyu Wang.
\newblock Defense against adversarial attacks using feature scattering-based
  adversarial training.
\newblock In {\em NIPS}, 2019.

\bibitem{semseg2019}
Hengshuang Zhao.
\newblock semseg.
\newblock \url{https://github.com/hszhao/semseg}, 2019.

\bibitem{zhao2017pyramid}
Hengshuang Zhao, Jianping Shi, Xiaojuan Qi, Xiaogang Wang, and Jiaya Jia.
\newblock Pyramid scene parsing network.
\newblock In {\em CVPR}, 2017.

\end{thebibliography}
}

\end{document}